\documentclass[runningheads]{llncs}
\usepackage[T1]{fontenc}
\usepackage{graphicx,verbatim}
\usepackage{booktabs}
\usepackage{amsmath}
\usepackage[table]{xcolor}
\usepackage{amssymb}
\usepackage{multirow}
\usepackage{booktabs,multirow,makecell,array}
\usepackage{mathtools}
\usepackage{bm}
\usepackage{algorithm}
\usepackage{algpseudocode}
\usepackage{cite}
\usepackage[colorlinks,
            linkcolor=blue,
            anchorcolor=blue,
            citecolor=blue]{hyperref}

\begin{document}

\title{CLoE: Expert Consistency Learning for Robust Missing Modality Segmentation}
%

\author{Xinyu Tong\textsuperscript{1}*, Meihua Zhou\textsuperscript{1}*†, Bowu Fan\textsuperscript{2}, and Haitao Li\textsuperscript{3}†}  

\authorrunning{X. Tong, M. Zhou et al.}
\institute{\textsuperscript{1}University of Chinese Academy Sciences \\ \textsuperscript{2} School of Medicine, Southeast University  \\
\textsuperscript{3} Zhejiang University \\
    \email{\{tongxinyu25,zhoumeihua25\}@mails.ucas.ac.cn, lihaitao@zju.edu.cn}}

\maketitle 
\begin{abstract}
Multimodal medical image segmentation frequently suffers from missing modalities during inference, causing inconsistent outputs across modality-specific experts and unstable feature fusion, especially for tiny foreground lesions. We propose Consistency Learning of Experts (CLoE), a consistency-driven segmentation framework that delivers competitive performance under both complete and incomplete multimodal inputs. CLoE builds robustness via dual decision-level consistency constraints: Modality Expert Consistency aligns global predictions to suppress case-wise drift with partial inputs, while Region Expert Consistency regularizes uniform outputs on clinically critical lesions to avoid background-biased optimization. A lightweight gating network converts cross-modal consistency scores into adaptive reliability weights to recalibrate features before fusion. Validated on BraTS-2020 and BraTS-Africa, CLoE outperforms excellent methods for missing-modality segmentation. Our consistency-guided gated fusion mitigates widespread modality scarcity and cross-site domain shift in resource-limited African healthcare, enabling reliable tumor segmentation without full multimodal scans. This framework shows promising potential for real-world medical AI deployment across diverse African clinical sites and provides a valuable paradigm for developing modality-robust segmentation models adapted to incomplete clinical multimodal data.


\end{abstract}
\section{Introduction}
Multimodal MRI segmentation is often developed assuming complete modality availability \cite{Charron2018,Qin2025BTSegDiff}. In clinical settings, missing sequences, protocol variation, and quality issues are common, so deployed systems must operate on arbitrary modality subsets while maintaining strong full-modality performance \cite{Liu2023M3AE,Liu2025Federated}. The challenge is not only reduced information but also unstable decisions \cite{Chen2025Fusion,Qin2025Thesis}. Modality-specific predictors can disagree, and fixed-weight fusion or unconstrained attention may amplify these discrepancies, especially in small yet critical regions. 

Encoder-decoder backbones such as U-Net \cite{unet} and V-Net \cite{vnet} remain the workhorse for volumetric segmentation, yet performance drops sharply when modalities are missing. This has motivated remedies that include generative synthesis with GAN models or Pix2Pix style translation \cite{gan_med,pix2pix}, arithmetic fusion such as HeMIS \cite{hemis}, and latent representation learning such as DC-Seg \cite{dcseg}. Adaptive fusion further reweights features through attention, including SE \cite{senet}, CBAM \cite{cbam}, and more dynamic designs \cite{contrastaware,damnet}, while RFNet exploits region aware priors on modality sensitivity \cite{rfnet}. Despite these advances, magnitude based attention can become uninformative when missing modalities are represented as zero tensors, and spatial priors are passive because they specify where to look rather than how reliable the available experts are. Consistency learning, effective in semi supervised settings such as Mean Teacher and extended via multimodal distillation and curriculum strategies \cite{DAGCL,meanteacher,dclse}, is also challenged by background dominance in volumetric MRI, where global agreement can be achieved without aligning small tumor subregions. Consequently, many approaches still rely on modality dropout or input synthesis \cite{Wang2023SharedSpecific}, improving average robustness but lacking an explicit mechanism to decide which modality expert should be trusted for a given case and region, so unequal evidence can induce expert drift that fusion propagates into segmentation errors.
\\ 
\indent We treat missing modality robustness as a decision level consistency problem and propose Consistency Learning of Experts (CLoE), which reduces expert disagreement during training and downweights unreliable experts during fusion by converting inter expert agreement into a reliability signal. CLoE introduces Expert Consistency Learning (ECL) with Modality Expert Consistency (MEC) to enforce global agreement and suppress case wise drift under missing inputs, and Region Expert Consistency (REC) to emphasize agreement on foreground structures where errors matter most, while a lightweight gating network maps MEC and REC scores to modality weights for feature recalibration before fusion without complicating inference. 
We benchmark CLoE on BraTS2020 and BraTS-Africa. While BraTS2020 is a well-established missing-modality dataset with clear lesion contrast, \textit{\textbf{BraTS-Africa is acquired from low-performance scanners in sub-Saharan African clinics}. Its faint tissue differentiation and late-stage glioma lesions introduce severe boundary ambiguity, challenging model generalization under degraded imaging conditions and limited training data.}
\\
\indent Our contributions are threefold. (i) We formulate missing-modality robustness as decision-level expert inconsistency control. (ii) We propose ECL with two complementary measurements: MEC for global agreement and REC for foreground-critical agreement. (iii) We introduce a consistency-driven gating network that converts MEC/REC into modality reliability weights for weighted fusion, improving incomplete-modality performance while preserving full-modality accuracy. (iv) CLoE is \textbf{particularly effective} on low-quality and rare data.

\begin{figure*}[t]
    \centering
    \includegraphics[width=\textwidth]{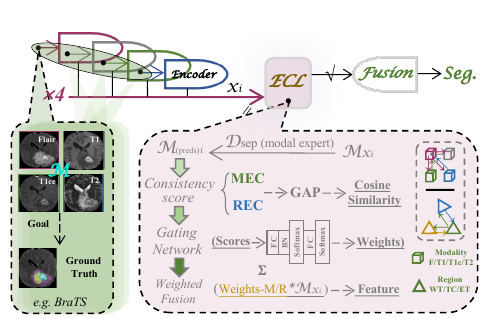}
    \caption{Overview of our proposed CLoE. This framework first extracts modality-specific features \textbf{\textit{M}x} from input volumes, undergoes regularization \textbf{\textit{M}pred}, and utilizes the ECL module to compute its Modality and Region Expert Consistency scores. Finally, it derives modality reliability weights via a lightweight gating network and performs consistency-driven weighted fusion to achieve robust missing modal segmentation.}
    \label{fig:main}
\end{figure*}

\section{Method}
We study multi-modal MRI (e.g., brain tumor) segmentation when a subset of modalities may be missing at inference time. Let $\mathcal{D}=\{(x_n,y_n)\}_{n=1}^{N}$ be the training set. Each sample contains $M$ modalities $x=\{x^{(m)}\}_{m=1}^{M}$ and a pixel-wise segmentation label $y\in\{0,1\}^{H\times W\times C}$ for $C$ classes. Modality availability is represented by a binary vector $s\in\{0,1\}^{M}$, defining the available set $\mathcal{A}=\{m\mid s_m=1\}$ and the effective inputs $\tilde{x}=\{\tilde{x}^{(m)}\}_{m=1}^{M}$ where $\tilde{x}^{(m)}=s_m x^{(m)}$. Our goal is to learn a mapping that produces a robust distribution $p\in[0,1]^{H\times W\times C}$ under arbitrary availability patterns $s$.

\subsection{Overview of Method}

As shown in Fig.~\ref{fig:main}, CLoE consists of parallel modality encoders, a consistency-driven gating module, and a shared fusion decoder. Each encoder $\Phi_m$ maps its input $\tilde{x}^{(m)}$ to multi-scale features $\{f^{(m)}_\ell\}_{\ell=1}^{L}$. A weight-shared expert decoder $D^{\mathrm{sep}}$ processes these features to yield individual expert predictions:
\begin{equation}
p^{(m)} = D^{\mathrm{sep}}(f^{(m)}_L), \qquad p^{(m)}\in[0,1]^{H\times W\times C}.
\end{equation}
A dynamic gating mechanism then aggregates the multi-scale features into a unified representation $F^{\mathrm{fuse}}=\{f_\ell\}_{\ell=1}^{L}$, which is processed by the fusion decoder $D^{\mathrm{fuse}}$ to produce the final mask $p=D^{\mathrm{fuse}}(f_L)$. 
The core challenge under partial observation is the decision instability: individual experts often yield conflicting predictions, and naive fusion amplifies this disagreement. CLoE explicitly tackles this by enforcing mutual agreement in the prediction space and utilizing this consistent signal to gate the feature fusion.

\subsection{Expert Consistency Measurement}
\label{sec:ecl}
To quantify agreement, we impose consistency constraints on the expert probability maps. Let $\mathbf{p}^{(m)}\in\mathbb{R}^{N}$ be the vectorized form of $p^{(m)}$ with $N=HWC$. We measure similarity using the cosine function:
\begin{equation}
\mathcal{S}(\mathbf{u},\mathbf{v})=\frac{\mathbf{u}^{\top}\mathbf{v}}{\|\mathbf{u}\|_2\|\mathbf{v}\|_2+\epsilon}.
\end{equation}
We define the set of available expert pairs as $\mathcal{P}=\{(a,b)\mid a,b\in\mathcal{A},\,a<b\}$.

\textbf{Modality Expert Consistency (MEC).}
MEC enforces global distribution alignment among the available experts:
\begin{equation}
\mathcal{L}_{\mathrm{MEC}}
=\frac{1}{|\mathcal{P}|}\sum_{(a,b)\in\mathcal{P}}\Bigl(1-\mathcal{S}(\mathbf{p}^{(a)},\mathbf{p}^{(b)})\Bigr).
\label{eq:mec}
\end{equation}

\textbf{Region Expert Consistency (REC).}
Since global agreement is often dominated by background pixels, we introduce a probabilistic region map $r\in[0,1]^{H\times W}$ to emphasize task-relevant structures. We compute $r$ from the aggregated shallow features of available experts via a lightweight projection head $\pi(\cdot)$:
\begin{equation}
r=\sigma\!\left(\pi\!\left(\frac{1}{|\mathcal{A}|}\sum_{m\in\mathcal{A}} f^{(m)}_{1}\right)\right),
\label{eq:region}
\end{equation}
where $\sigma(\cdot)$ is the sigmoid function. By broadcasting $r$ across classes, we define the region-weighted prediction vector $\mathbf{p}^{(m)}_{r}=\mathrm{vec}(r\odot p^{(m)})$. The REC constraint is formulated as:
\begin{equation}
\mathcal{L}_{\mathrm{REC}}
=\frac{1}{|\mathcal{P}|}\sum_{(a,b)\in\mathcal{P}}\Bigl(1-\mathcal{S}(\mathbf{p}^{(a)}_{r},\mathbf{p}^{(b)}_{r})\Bigr).
\label{eq:rec}
\end{equation}

\subsection{Consistency-Driven Dynamic Gating}
We utilize the calculated consistency not only as a constraint but also as a reliability cue for feature fusion. For each available expert $m\in\mathcal{A}$, we compute its global and regional consistency scores:
\begin{equation}
u_m=\frac{1}{|\mathcal{A}|-1}\sum_{k\in\mathcal{A}\setminus\{m\}}\mathcal{S}(\mathbf{p}^{(m)},\mathbf{p}^{(k)}),
\qquad
v_m=\frac{1}{|\mathcal{A}|-1}\sum_{k\in\mathcal{A}\setminus\{m\}}\mathcal{S}(\mathbf{p}^{(m)}_{r},\mathbf{p}^{(k)}_{r}).
\label{eq:scores}
\end{equation}
A lightweight gating network $\mathcal{G}$ maps the tuple $(u_m,v_m)$ to a reliability logit $g_m$. These logits are normalized via softmax over the available experts to obtain the fusion weights $w\in[0,1]^M$:
\begin{equation}
w_m=
\begin{cases}
\displaystyle \frac{\exp(g_m)}{\sum_{k\in\mathcal{A}}\exp(g_k)}, & m\in\mathcal{A},\\[6pt]
0, & m\notin\mathcal{A},
\end{cases}
\qquad g_m=\mathcal{G}(u_m,v_m).
\label{eq:gating}
\end{equation}
The multi-scale features are then adaptively fused via $f_\ell=\sum_{m=1}^{M} w_m \odot f^{(m)}_\ell$. This inherently suppresses experts that deviate from the consistency, stabilizing the representation under missing modalities.

\subsection{Learning Process}
Unlike existing methods (e.g., DC-Seg~\cite{dcseg}) that rely on latent disentanglement purely for representation accuracy, our framework explicitly tackles the vulnerability of missing modalities through a unified \textit{robustness paradigm}. The overall learning objective comprises three distinct components.

\textbf{Robust Expert Consistency Learning ($\mathcal{L}_{\mathrm{ECL}}$).}
To achieve true robustness, modality experts must possess foundational discriminative capabilities while simultaneously reaching mutual consistency. We integrate the independent expert supervision with our proposed consistency constraints into a unified ECL objective:
\begin{equation}
\mathcal{L}_{\mathrm{ECL}} = \underbrace{\sum_{m \in \mathcal{A}} \Big( \mathcal{L}_{\mathrm{WCE}}(p^{(m)}, y) + \mathcal{L}_{\mathrm{DL}}(p^{(m)}, y) \Big)}_{\text{Individual Expert Supervision}} + \eta \underbrace{\Big( \mathcal{L}_{\mathrm{MEC}} + \lambda_{\mathrm{rec}} \mathcal{L}_{\mathrm{REC}} \Big)}_{\text{Mutual Consistecy}},
\end{equation}
where the first term anchors the basic segmentation ability of each expert, serving as a prerequisite for the meaningful global ($\mathcal{L}_{\mathrm{MEC}}$) and regional ($\mathcal{L}_{\mathrm{REC}}$) consistency derived in Sec.~\ref{sec:ecl}. $\eta$ balances the two components.

\textbf{Fusion Segmentation Loss ($\mathcal{L}_{\mathrm{seg}}$).}
The fused feature representation, dynamically aggregated via our gating module, is supervised by the main task loss to align the consistency-driven prediction with the ground truth:
\begin{equation}
\mathcal{L}_{\mathrm{seg}} = \mathcal{L}_{\mathrm{WCE}}(p, y) + \mathcal{L}_{\mathrm{DL}}(p, y).
\end{equation}

\textbf{Contrastive Representation Loss ($\mathcal{L}_{\mathrm{contrast}}$).}
To explicitly enhance representation accuracy by disentangling the latent space, we introduce a composite contrastive objective~\cite{dcseg} that aligns anatomical content, clusters modality styles, and enforces generative validity:
\begin{equation}
\begin{aligned}
\mathcal{L}_{\mathrm{contrast}} = & \sum_{i,j} \log \Big(1 + \exp\big(-\delta_{ij} \tau \text{SSIM}(a_i, a_j)\big)\Big) \\
& + \sum_{i,j} \log \Big(1 + \exp\big(-\delta_{ij} \tau \cos(m_i, m_j)\big)\Big) \\
& + \gamma \Big( \|x - \hat{x}\|_1 + \mathcal{D}_{\mathrm{KL}}(z \| \mathcal{N}(0,I)) \Big).
\end{aligned}
\label{eq:contrast}
\end{equation}

\textbf{Total Objective.}
The overall objective of our CLoE framework seamlessly balances task accuracy, consistency-driven robustness, and representational disentanglement:
\begin{equation}
\mathcal{L}_{\mathrm{total}} = \mathcal{L}_{\mathrm{seg}} + \alpha \mathcal{L}_{\mathrm{ECL}} + \beta \mathcal{L}_{\mathrm{contrast}},
\label{eq:total}
\end{equation}
where $\alpha$ and $\beta$ are hyperparameters.

\begin{algorithm}[t]
\small
\caption{CLoE Core: Expert Consistency and Consistency-Driven Gating}
\label{alg:cloe_core}
\begin{algorithmic}[1]
\Require Expert features $\{f^{(m)}\}$, Latent codes $\{a, m\}$, Ground truth $y$
\Ensure Total loss $\mathcal{L}_{\mathrm{total}}$ and Gating weights $w$

\State $\mathcal{A} \leftarrow \{m \mid s_m=1\}$; $\mathcal{P}\leftarrow\{(a,b)\mid a,b\in\mathcal{A},\,a<b\}$

\Statex \textbf{Step 1: Robust Expert Consistency Learning ($\mathcal{L}_{\mathrm{ECL}}$)}
\State Predict individual masks: $p^{(m)} \leftarrow D^{\mathrm{sep}}(f^{(m)}_L)$ for all $m \in \mathcal{A}$
\State Compute Individual Supervision via $\sum(\mathcal{L}_{\mathrm{WCE}} + \mathcal{L}_{\mathrm{DL}})$ on $\{p^{(m)}\}$
\State Vectorize maps: $\mathbf{p}^{(m)} \leftarrow \mathrm{vec}(p^{(m)})$; Compute Region Map $r$
\State Compute Mutual Consistency terms: $\mathcal{L}_{\mathrm{MEC}}$ and $\mathcal{L}_{\mathrm{REC}}$
\State $\mathcal{L}_{\mathrm{ECL}} \leftarrow \text{Individual Supervision} + \eta (\mathcal{L}_{\mathrm{MEC}} + \lambda_{\mathrm{rec}}\,\mathcal{L}_{\mathrm{REC}})$

\Statex \textbf{Step 2: Contrastive Representation Loss ($\mathcal{L}_{\mathrm{contrast}}$)}
\State Compute alignment terms: SSIM (Content) and Cosine (Style)
\State Compute generative terms: $L_1$ Reconstruction and KL Divergence
\State $\mathcal{L}_{\mathrm{contrast}} \leftarrow \text{Sum of terms defined in Eq.}~\eqref{eq:contrast}$

\Statex \textbf{Step 3: Fusion \& Total Objective ($\mathcal{L}_{\mathrm{seg}}$ \& $\mathcal{L}_{\mathrm{total}}$)}
\State Calculate gating weights $w$ based on consistency scores from Step 1
\State Fuse features: $f^{\mathrm{fuse}} \leftarrow \sum_{m \in \mathcal{A}} w_m f^{(m)}$
\State Predict $p \leftarrow D^{\mathrm{fuse}}(f^{\mathrm{fuse}})$ and compute $\mathcal{L}_{\mathrm{seg}}$ via $(\mathcal{L}_{\mathrm{WCE}} + \mathcal{L}_{\mathrm{DL}})$
\State Update via $\mathcal{L}_{\mathrm{total}} = \mathcal{L}_{\mathrm{seg}} + \alpha \mathcal{L}_{\mathrm{ECL}} + \beta \mathcal{L}_{\mathrm{contrast}}$

\end{algorithmic}
\end{algorithm}

\section{Experiments and Results}

\textbf{Datasets and Implementation Details.} All experiments are conducted on two publicly available medical image segmentation benchmarks: BraTS 2020 and BraTS-Africa dataset from African medical centers using lower-field MRI systems that have lower image contrast and resolution.\cite{africa}. BraTS 2020 includes \textbf{369} multi-contrast MRI scans covering T1, T1c, T2, and FLAIR modalities, with annotations for three tumor subregions: whole tumor, tumor core, and enhancing tumor. The dataset collectors preprocessed all MRI images via skull stripping, co-registration, and resampling to 1 mm³ isotropic resolution. For model training, 112 × 112 × 112 3D patches were randomly cropped from preprocessed volumes as network inputs. BraTS-Africa dataset contains \textbf{60} cases multimodal MR volumes with four registered MRI modalities (T1n, T2w, T2f, and T1c)(Structure: NC, Oedema and ET), Data preprocessing procedures align with those of BRATS 2020. Data augmentation includes random flipping, random cropping, and intensity variation. Experiments (CLoE and DC-Seg\cite{dcseg}) employed the Adam optimizer for training with an initial learning rate of 0.0002, weight decay of 0.0001, and 500 epochs of training with batch size of 1.

\begin{table}[t]
\centering
\caption{Performance comparison (Dice\%) with SOTA methods
on BraTS 2020.}
\tiny
\begin{tabular}{l|l|llll|llllll|lllll|l}
\ \#              & Methods   & T2     & T1c   & T1     & F    & \begin{tabular}[c]{@{}l@{}}T2\\ T1c\end{tabular} & \begin{tabular}[c]{@{}l@{}}T1c\\ T1\end{tabular} & \begin{tabular}[c]{@{}l@{}}T1\\ F\end{tabular} & \begin{tabular}[c]{@{}l@{}}T2\\ T1\end{tabular} & \begin{tabular}[c]{@{}l@{}}T2\\ F\end{tabular} & \begin{tabular}[c]{@{}l@{}}T1c\\ F\end{tabular} & \begin{tabular}[c]{@{}l@{}}$\sim$\\ T2\end{tabular} & \begin{tabular}[c]{@{}l@{}}$\sim$\\ T1c\end{tabular}  & \begin{tabular}[c]{@{}l@{}}$\sim$\\ T1\end{tabular}  & \begin{tabular}[c]{@{}l@{}}$\sim$\\ F\end{tabular} & Full  & Avg   \\ \cmidrule{1-18}
\multirow{6}{*}{WT} & HeMIS     & 79.85  & 64.58 & 63.01  & 52.29 & 84.45  & 72.50  & 65.29 & 82.31  & 81.56 & 69.37  & 73.31    & 83.03     & 84.64    & 85.19   & 85.19 & 75.10 \\
                    & RobustSeg & 82.20  & 71.39 & 71.41  & 82.87 & 85.97  & 76.84  & 88.10 & 85.53  & 88.09 & 87.33  & 88.87    & 89.24     & 88.68    & 86.63   & 89.47 & 84.17 \\
                    & RFNet     & 86.05  & 76.77 & 77.16  & 87.32 & 87.74  & 81.12  & 89.73 & 87.73  & 89.87 & 89.89  & 90.69    & 90.60     & 90.68    & 88.25   & 91.11 & 86.98 \\
                    & M³AE      & 86.10  & 78.90 & 79.00  & 88.00 & 87.10  & 80.10  & 89.60 & 87.30  & 90.10 & 89.50  & 89.60    & 90.20     & 90.50    & 87.40   & 90.40 & 86.90 \\
                    & DC-Seg    & 86.67  & 79.36 & 79.31  & 88.67 & 88.57  & 83.03  & 90.18 & 88.29  & 90.41 & 90.44  & \cellcolor{gray!20}90.98    & 90.80     & 91.09    & 88.99   & 91.22 & 87.87 \\
                    & CLoE      & \cellcolor{gray!20}87.19  & \cellcolor{gray!20}80.08 & \cellcolor{gray!20}79.96  & \cellcolor{gray!20}88.90 & \cellcolor{gray!20}88.63  & \cellcolor{gray!20}83.57  & \cellcolor{gray!20}90.32 & \cellcolor{gray!20}88.42  & \cellcolor{gray!20}90.47 & \cellcolor{gray!20}90.49  & 90.92    & \cellcolor{gray!20}90.87     & \cellcolor{gray!20}91.18    & \cellcolor{gray!20}89.06   & \cellcolor{gray!20}91.30 & \cellcolor{gray!20}88.09 \\ \cmidrule{1-18}
\multirow{6}{*}{TC} & HeMIS     & 54.22  & 69.41 & 42.42  & 24.97 & 77.60  & 75.59  & 41.58 & 56.38  & 55.89 & 70.86  & 75.07    & 57.40     & 77.69    & 79.05   & 78.58 & 65.45 \\
                    & RobustSeg & 61.88  & 76.68 & 54.30  & 60.72 & 82.44  & 80.28  & 68.18 & 66.46  & 68.20 & 81.85  & 82.76    & 70.46     & 81.89    & 82.85   & 82.87 & 73.45 \\
                    & RFNet     & 71.02  & 81.51 & 66.02  & 69.19 & 83.45  & 83.40  & 73.07 & 73.13  & 74.14 & 84.65  & 85.07    & 75.19     & 84.97    & 83.47   & 85.21 & 78.23 \\
                    & M³AE      & 71.80  & 83.60 & \cellcolor{gray!20}69.40  & 68.70 & 85.60  & 83.80  & 72.80 & 72.90  & 74.30 & 85.50  & 85.60    & 74.40     & 85.80    & 85.80   & 86.20 & 79.10 \\
                    & DC-Seg    & \cellcolor{gray!20}71.91  & 85.34 & 69.01  & 71.22 & 86.30  & 86.81  & 75.13 & 73.56  & 74.41 & 85.89  & 86.49    & 75.34     & 85.90    & 86.55   & 86.19 & 80.00 \\
                    & CLoE      & 71.36  & \cellcolor{gray!20}85.41 & \cellcolor{gray!20}69.40  & \cellcolor{gray!20}71.34 & \cellcolor{gray!20}86.68  & \cellcolor{gray!20}87.00  & \cellcolor{gray!20}75.49 & \cellcolor{gray!20}73.61  & \cellcolor{gray!20}74.48 & \cellcolor{gray!20}85.91  & \cellcolor{gray!20}86.60    & \cellcolor{gray!20}75.36     & \cellcolor{gray!20}86.74    & \cellcolor{gray!20}87.14   & \cellcolor{gray!20}87.06 & \cellcolor{gray!20}80.23 \\ \cmidrule{1-18}
\multirow{6}{*}{ET} & HeMIS     & 31.43  & 63.24 & 16.53  & 9.00  & 70.30  & 70.71  & 13.99 & 28.58  & 28.91 & 68.31  & 70.80    & 29.53     & 71.36    & 71.67   & 71.49 & 47.73 \\
                    & RobustSeg & 36.46  & 67.91 & 28.99  & 34.68 & 71.42  & 70.11  & 39.67 & 39.92  & 42.19 & 70.78  & 71.77    & 43.90     & 71.17    & 71.89   & 71.52 & 55.49 \\
                    & RFNet     & 46.29  & 74.85 & 37.30  & 38.15 & 75.93  & 78.01  & 40.98 & 45.65  & 49.32 & 76.67  & 76.81    & 49.92     & 77.12    & 76.99   & 78.00 & 61.47 \\
                    & M³AE      & \cellcolor{gray!20}47.10  & 73.60 & 40.40  & 40.20 & 76.00  & 75.30  & 43.70 & \cellcolor{gray!20}48.70  & 47.10 & 75.90  & 76.30    & 48.20     & 77.40    & 78.00   & 77.50 & 61.70 \\
                    & DC-Seg    & 44.34  & 77.57 & \cellcolor{gray!20}40.83  & 42.56 & 79.35  & 81.54  & 44.98 & 47.87  & 50.48 & 78.46  & 80.79    & 49.38     & 80.19    & 81.97   & 80.03 & 64.02 \\
                    & CLoE      & 46.61 & \cellcolor{gray!20}78.70 & 39.29 & \cellcolor{gray!20}44.35 & \cellcolor{gray!20}79.97  & \cellcolor{gray!20}81.12  & \cellcolor{gray!20}47.45 & 48.69 & \cellcolor{gray!20}51.35 & \cellcolor{gray!20}79.86  & \cellcolor{gray!20}81.55    & \cellcolor{gray!20}52.44     & \cellcolor{gray!20}81.48    & \cellcolor{gray!20}81.61   & \cellcolor{gray!20}81.46 & \cellcolor{gray!20}65.06
\end{tabular}
\label{tab:compare1}
\end{table}

\textbf{Compare Experimental Models.} We compare our proposed CLoE framework with several state-of-the-art(SOTA) methods, including HeMIS \cite{hemis}, RobustSeg \cite{robustseg}, RFNet \cite{rfnet} and M³AE \cite{m3ae}. We follow the same data split as these prior works and directly report their published results to ensure a fair comparison. 

As shown in Table~\ref{tab:compare1}, among the 15 missing-modality combinations, our CLoE method outperforms all competing approaches in the whole tumor (WT) segmentation task, achieving an average Dice coefficient of 88.09\%. This result surpasses both M³AE (86.90\%) and DC-Seg (87.54\%). In addition, for tumor core (TC) segmentation, CLoE yields an average Dice of 80.23\%, which exceeds DC-Seg (79.63\%) and M³AE (79.10\%), further demonstrating its strong robustness against missing-modality inputs. Furthermore, in the enhancing tumor (ET) segmentation task, CLoE achieves an average Dice score of 65.06\%, which is comparable to DC-Seg (65.00\%) and considerably higher than M³AE (61.70\%). Notably, our unified single-model framework delivers superior segmentation accuracy while maintaining strong robustness, outperforming large pre-trained models such as M³AE and specialized methods including DC-Seg, without the need for training separate models for each modality combination. 

\begin{figure*}[t]
    \centering
    \includegraphics[width=\textwidth]{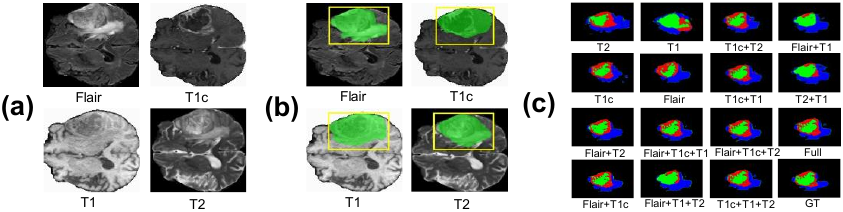}
    \caption{(a) Visualization of the input modalities. (b) General MedSAM model prediction with bounding box prompt. (c) CLoE predicted segmentation maps.}
    \label{fig:vis}
\end{figure*}

Fig.~\ref{fig:vis} demonstrates that our method can effectively segment brain tumors in scenarios with missing data across different modalities. Additionally, we evaluate the general medical foundation model MedSAM\cite{medsam} on the full-modality setting. Even with bounding box prompts, MedSAM cannot generate clear and accurate tumor boundaries, which demonstrates the value of our dedicated multimodal segmentation framework, especially for incomplete-modality cases.

Table~\ref{tab:compare2} presents the results for the BraTS-Africa dataset on \textbf{ET}. \textbf{Modalities (c:T1c, n:T1n, f:T2f, w:T2w)}. Compared with RFNet, Our CLoE shows strong robustness to modality changes. CLoE improves the average Dice by 11.33\% over RFNet. (\textbf{\textit{In particular, the average: DC-Seg is only 40.3\%}})

\begin{table}[htbp]
  \centering
  \footnotesize 
  \begin{minipage}{0.40\textwidth}
  \setlength{\tabcolsep}{1.0pt}
    \centering
    \footnotesize 
    \caption{*Performance of SOTA vs CLoE on BraTS-Africa dataset.}
    \renewcommand{\arraystretch}{1.2} 
    \begin{tabular}{llllllll}
\ \ \textbf{c}& \ \ \textbf{n}& \ \ \textbf{f}& \ \textbf{w}& \textbf{c,n}& \ \textbf{c,f}& \textbf{c,w}& \ \textbf{n,f}  \\ 

46.2& 17.5& 26.8& 19.1& 54.2& 50.0& 50.8& 31.2    \\
      \cmidrule{1-8}
\textbf{n,w}& \ \textbf{f,w}& \ \ \textbf{\~c}& \ \ \textbf{\~n}& \ \ \textbf{\~f}& \ \textbf{\~w}& \textbf{Full}& \textbf{Avg}   \\ 

24.5& 36.6& 35.6& 54.4& 57.7& 57.4& 60.0& 41.5    \\
      \toprule
\ \ \textbf{c}& \ \ \textbf{n}& \ \ \textbf{f}& \ \textbf{w}& \textbf{c,n}& \ \textbf{c,f}& \textbf{c,w}& \ \textbf{n,f}  \\ 

\cellcolor{gray!20}54.6& \cellcolor{gray!20}20.6& \cellcolor{gray!20}28.3& \cellcolor{gray!20}26.9& \cellcolor{gray!20}62.5& \cellcolor{gray!20}53.8& \cellcolor{gray!20}58.2& \cellcolor{gray!20}33.7    \\
      \cmidrule{1-8}
\textbf{n,w}& \ \textbf{f,w}& \ \ \textbf{\~c}& \ \ \textbf{\~n}& \ \ \textbf{\~f}& \ \textbf{\~w}& \textbf{Full}& \textbf{Avg}   \\ 

\cellcolor{gray!20}31.3& \cellcolor{gray!20}35.4& \cellcolor{gray!20}37.5& \cellcolor{gray!20}57.9& \cellcolor{gray!20}64.3& \cellcolor{gray!20}63.1& \cellcolor{gray!20}65.0& \cellcolor{gray!20}46.2    \\
    \end{tabular}
    \label{tab:compare2}
  \end{minipage}
  \hfill
  \begin{minipage}{0.51\textwidth}
    \centering
    \footnotesize 
    \caption{Ablation of CLoE on BraTS2020. (\textbf{LC}: L-contrast. \textbf{$^\dagger$}: w/ Gating Network. \textbf{$^\ddagger$}: w/ Gating Network \& Weight Fusion.)}
    \begin{tabular}{l|lll}
    \textbf{Methods} & \textbf{WT} & \ \textbf{TC} & \ \textbf{ET}      \\ 
      \cmidrule{1-4}
    Baseline \cite{rfnet} & 86.98 & 78.23 & 61.47  \\
    (+ LC) \cite{dcseg} & 87.87 & 80.00 & 64.02  \\
      \cmidrule{1-4}
    + LC + MEC & 87.91 & 80.02 & 64.11  \\
    + LC + MEC + REC & 88.01 & 80.10 & 64.65  \\
      $^\dagger$(+ LC + MEC + REC) & 88.04 & 80.14 & 64.87  \\
      \cellcolor{gray!20}$^\ddagger$(+ LC + MEC + REC) & \cellcolor{gray!20}88.09 & \cellcolor{gray!20}80.23 & \cellcolor{gray!20}65.06 \\
    \cmidrule{1-4}
     $^\ddagger$(+ MEC + REC) & 87.72 & 79.88 & 64.13  \\
    \end{tabular}
    \label{tab:ablation}
  \end{minipage}
\end{table}

\textbf{Ablation Study.} Table~\ref{tab:ablation} compares CLoE with RFNet as the baseline. L-contrast (LC) improves accuracy through cross-modal alignment (DC-Seg). MEC yields a slight improvement, while REC further enhances performance. Adding only the gating network ($^\dagger$) yields negligible improvements, whereas combining the gating network with adaptive weighted fusion ($^\ddagger$) achieves the best Dice scores, which are 1.11\%, 2.00\%, and 3.59\% higher than the baseline model, respectively. Finally, when only LC is removed, its level is comparable to (+ LC).

\section{Conclusion}
We presented CLoE for multimodal MRI segmentation with missing modalities. CLoE frames robustness as decision-level consistency and stabilizes modality experts via dual-branch Expert Consistency Learning, where Modality Expert Consistency enforces global agreement to reduce case-wise drift and Region Expert Consistency emphasizes foreground agreement to mitigate background dominance. A lightweight gating network converts agreement into reliability weights to recalibrate modality features before fusion without increasing inference complexity. Experiments on BraTS 2020 and BraTS-Africa show consistent gains across missing-modality patterns while preserving full-modality performance, and ablations confirm the complementary effects of regional consistency and consistency-guided fusion on clinically critical subregions.


%
%
%
%

\end{document}